\pdfoutput=1
\documentclass{bmvc2k}

\title{Adaptation of Distinct Semantics for Uncertain Areas in Polyp Segmentation}

\addauthor{Quang Vinh Nguyen}{vinhbn28@jnu.ac.kr}{1}
\addauthor{Van Thong Huynh}{vthuynh@jnu.ac.kr}{1}
\addauthor{Soo-Hyung Kim}{shkim@jnu.ac.kr}{1}

\addinstitution{
 Department of AI Convergence\\
 Chonnam National University\\
 Gwangju, South Korea
}
\runninghead{Quang Vinh Nguyen, Van Thong Huynh, Soo-Hyung Kim}{ADSNet}

\begin{document}

\maketitle

\begin{abstract}
Colonoscopy is a common and practical method for detecting and treating polyps. Segmenting polyps from colonoscopy image is useful for diagnosis and surgery progress. Nevertheless, achieving excellent segmentation performance is still difficult because of polyp characteristics like shape, color, condition, and obvious non-distinction from the surrounding context. This work presents a new novel architecture namely Adaptation of Distinct Semantics for Uncertain Areas in Polyp Segmentation (ADSNet), which modifies misclassified details and recovers weak features having the ability to vanish and not be detected at the final stage. The architecture consists of a complementary trilateral decoder to produce an early global map. A  continuous attention module modifies semantics of high-level features to analyze two separate semantics of the early global map. The suggested method is experienced on polyp benchmarks in learning ability and generalization ability, experimental results demonstrate the great correction and recovery ability leading to better segmentation performance compared to the other state of the art in the polyp image segmentation task. Especially, the proposed architecture could be experimented flexibly for other CNN-based encoders, Transformer-based encoders, and decoder backbones. 
\end{abstract}

\section{Introduction}
\label{sec:intro}

Image segmentation is a major and significant topic in computer vision. This task classifies each pixel of the incoming image into predetermined classifications. Medical image segmentation is one of the highlight applications of segmentation techniques such as polyp segmentation, brain tumor segmentation, skin lesion segmentation, or lung segmentation. 

Due to its complexity, polyp segmentation has recently attracted a lot of interest. Polyps are abnormal tissue growth from the surface of internal organs and can be found in the colon, rectum, stomach, or even throat. In most cases, polyps are benign, which means that they do not indicate illness or maliciousness. However, since polyps are capable of developing into cancer, a long-term diagnostic is necessary to determine whether or not they have become malignant. Therefore, identifying polyps in the colonoscopy image is helpful in facilitating the early detection of polyp-related diseases. Colonoscopy is the primary method to locate and remove polyps, but this procedure requires a significant amount of time. In addition, there are still some challenges in a practical setting: common polyps can differ in size, color, and shape. Besides, polyps can develop haphazardly or densely in numerous sites, and may be challenging to distinguish them from surrounding tissues. This requires a reliable endoscopic polyps segmentation approach with high segmentation efficiency in difficult contexts.

\begin{figure}[htp]
\centering
 \includegraphics[scale=.65]{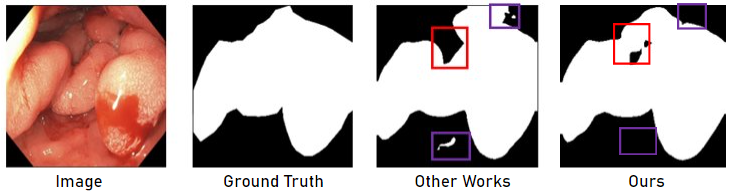}
\caption{Segmentation example of our model. Red boxes refer to uncertain areas. Purple boxes stand for noise details.}
\label{fig:refhinh1}
\end{figure}

In general, difficult details and hazy areas pose restrictions to medical image segmentation architectures. Previous deep learning-based works focused on the result at the final stage without paying attention to modifying and recovering these areas. Therefore, modern architecture systems struggle to identify and categorize challenging features and unclear areas, as shown in~\autoref{fig:refhinh1}. The area indicated in the red box of other works is fairly dark and separates from the nearby polyp items, in this case, the model could not identify sufficient desired objects. Similarly, areas in the purple box are confused with actual polyps, causing misclassification. Under our supervision, these areas might be correctable and recoverable, since they gradually weaken and are diluted with decoder progress. Motivated by this, we propose a new approach, Adaptation of Distinct Semantics for Uncertain Areas in Polyp Segmentation (ADSNet), that uses a complementary trilateral decoder to give a superior early global map. Proposing a new module: continuous attention, we fix inaccurate details and restore missed areas following two semantics: background semantic and object semantic. This strategy improves learning ability, generalization capability and overcomes the weaknesses of current models in challenging contexts. 

The following sections are organized as follows: In section II, we present outstanding methods for salient object detection and associated reports to the polyp segmentation task. Section III explores the ADSNet components, while section IV executes experiments to observe the performance of the proposed solution. In the final section, we summarize the entire contributions of the paper.

\section{Related Work}

\paragraph{Salient Object Detection.} SOD is a crucial preprocessing method for many computer vision applications including semantic segmentation, visual tracking, and image retrieval. Traditional SOD solutions are mainly based on heuristic priors (such as color, texture, and contrast) to construct saliency maps. With the advancement of Deep Neural Networks (DNNs), salient object detection (SOD)~\cite{borji2019salient}~\cite{wang2021salient} has made significant strides. First, several methods pay attention to improve the accuracy, Edge Guidance Network (EGNet)~\cite{zhao2019egnet} combines each path from the top-down stream for the salient object branch and the edge detection branch. BANet~\cite{su2019selectivity} employed side-out fusion and single-stream, respectively, for boundary and object branches. To forecast salient objects with complete structure and exquisite borders, AFNet~\cite{feng2019attentive} proposed a multi-scale attentive feedback model and Boundary-Enhanced Loss. Other approaches consider striking a balance between accuracy and efficiency, RAS~\cite{chen2018reverse} explicitly multiplies a reverse prediction region to acquire the remaining features for saliency refinement. CPD~\cite{wu2019cascaded} introduced a cascade partial decoder that uses an attention strategy to enhance high-level features while discarding shallow layer features for acceleration. ITSDNet~\cite{zhou2020interactive} proposed an interactive two-stream decoder to investigate several cues, such as saliency, contour, and their interaction. 

\paragraph{Polyp Segmentation.} With the development of deep learning techniques, the polyp segmentation task is explored strongly. Brandao et al.~\cite{brandao2017fully} proposed a fully convolutional neural network to gain great polyp segmentation performance. The first U-shape architecture Unet~\cite{ronneberger2015u} with two paths: an encoder to extract features and a decoder to educate the final map proved excellent performance in biomedical image segmentation. Motivated by the U-shape structure, several advanced versions were introduced. ResUnet++~\cite{jha2019resunet++} utilized the advantages of residual link, channel attention, and Atrous Spatial Pyramidal Pooling. Unet++~\cite{zhou2018unet++} kept rich information by dense concatenation via multi-stages. Several other variants use multi U-shape like DoubleUnet~\cite{jha2020doubleu}, and CUNet~\cite{tang2018cu} to achieve better feature representation. Newer methods focus on the relationship between area and boundary, SFANet~\cite{fang2019selective} considered area and border constraints. PraNet~\cite{fan2020pranet} applied Reverse Attention Module to correct the boundary area to boost the segmentation performance. Inspired by PraNet~\cite{fan2020pranet}, UACANet~\cite{kim2021uacanet} was designed with a new attention: parallel axial attention and augmenting uncertain area to model border information. SSFormer~\cite{wang2022stepwise} used a pyramid Transformer encoder to improve the generalization ability of models and propose a new decoder to emphasize local features. Another transformer-based architecture, Polyp-PVT~\cite{dong2021polyp} proposed a similarity aggregation module to extract local pixels and global semantic cues from the polyp area, effectively suppressing noise in features and significantly improving their expressive capabilities.

\section{Methodology}
The overall proposed architecture is visualized in~\autoref{fig:refhinh2}. The network extracts four levels of spatial features $\{f_i:i=1,...,4\}$. The complementary trilateral decoder~\cite{zhao2021complementary} constructs an early global map from encoder features. The continuous attention produces two distinct semantic masks from weak and strong regions before concatenating with the early global map to give a superior final mask. Each component will be explained in further detail in the following subsections.

\subsection{Encoder Backbone}\label{AA}
In computer vision tasks in general, the encoder component extracts features that bring important semantics for analyzing objects. Recent works often used CNN-based, Transformer-based encoders or combined both in several particular situations. CNN-based approaches do well in extracting local features by using the local kernel, Transformer-based architectures are effective in grasping global relationships, improving generalization ability, and multi-scale feature processing ability, but they need more data or a strong pre-trained. In a different approach without transformer, our work uses a CNN-based encoder to extract features and exploit the features obtained by analyzing separate semantics. This also brings many advances in generalization ability and multi-scale feature processing ability. In particular, we use Efficientnet-V2S~\cite{tan2019efficientnet} as an encoder backbone that produces multiple levels of spatial features $f_i \in R^{\frac{H}{2^{i+1}}*\frac{W}{2^{i+1}}*C_i}$ where $C_i \in \{256, 512, 1024, 2048\}$ and $i \in \{1,2,3,4\}$.

\subsection{Complementary Trilateral Decoder}\label{AA}
Deep features offer important information for locating objects, while shallow features have high resolution and do not store many valuable textures. Therefore, we focus on utilizing high-level features with no regard for low-level features. In this work, we introduce a Complementary Trilateral Decoder~\cite{zhao2021complementary} - a new SOTA decoder for silent object detection to address the loss of spatial structure, lack of boundary detail, and diluted semantic context. The early global map is obtained by $M = CTD(f_1,f_2,f_3,f_4).$ By estimating and using a decision parameter, we divide the early global map into two weak and strong regions to analyze separated semantics. The first component is strong ($S$) areas that hold clear object structure, and the second one is weak ($W$) areas that are determined by the decision parameter and refer to uncertain areas that could be reconstructed. 

\begin{figure}[htp]
    \begin{center}
     \includegraphics[width=.95\linewidth]{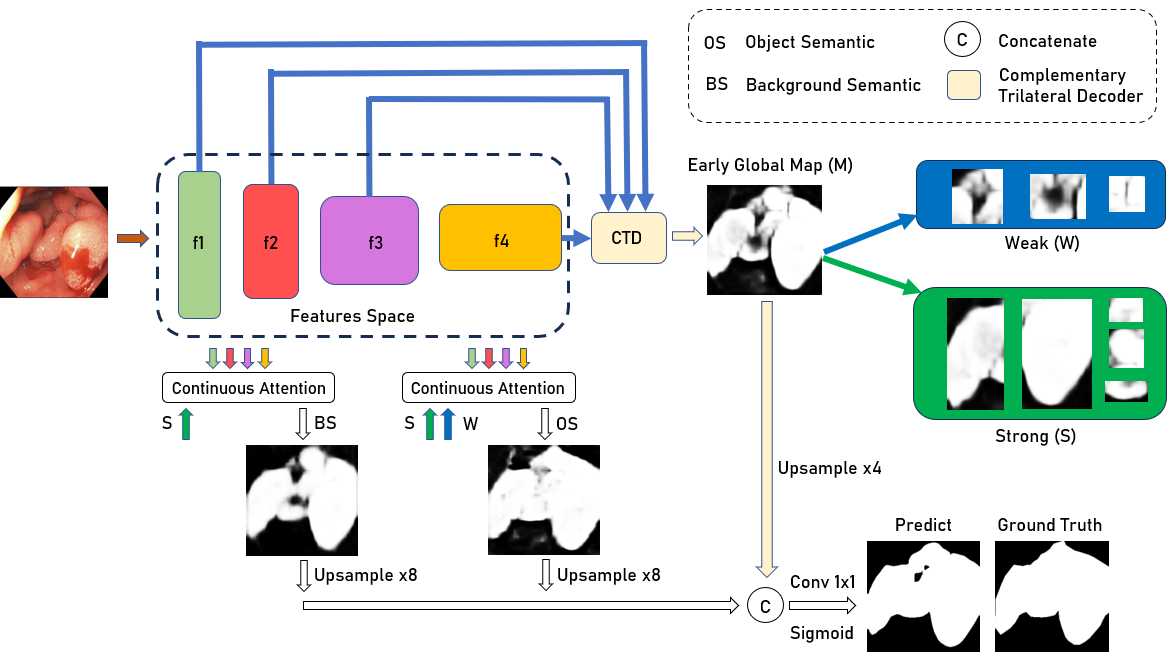}
     \end{center}
    \caption{The proposed ADSNet architecture.}
    \label{fig:refhinh2}
\end{figure}

\subsection{Continuous Attention for Background Semantic and Object Semantic}

The multi-scale representation can assist in perceiving multi-scale objects for SOD. Particularly, polyps frequently come in a variety of sizes, hence we suggest a unique structure called Progress Atrous Spatial Pyramidal Pooling (PASPP)~\cite{yan2020covid} to progressively capture multi-scale high-level features $\{f_2,f_3,f_4\}$. Lower-level features $\{f_1\}$ often hold rich detail information in appearance, we adapt Camouflage Identification Module (CIM)~\cite{dong2021polyp} to represent better texture and edge information. The CIM~\cite{dong2021polyp} is operated by the following two consecutive attention mechanisms:
\begin{align}
Att_c = \sigma(H_1(G_{max}(x))+H_2(G_{avg}(x))) \otimes x\\
Att_s = \sigma(Conv3x3(Cat(R_m(x),R_a(x))))\otimes x
\end{align}
\indent Where $x$ is the input feature. $G_{max}, G_{avg}$ are global max pooling and global average pooling functions, respectively. $H_1, H_2$ are two convolutional layers to reduce the dimension and recover the original dimension. $R_{m}, R_{a}$ are max pooling and average pooling following channel dimension. $Cat$ is concatenate operation, while $\sigma$ stands for Sigmoid function.
\begin{figure}[htp]
\begin{center}
 \includegraphics[width=.97\linewidth]{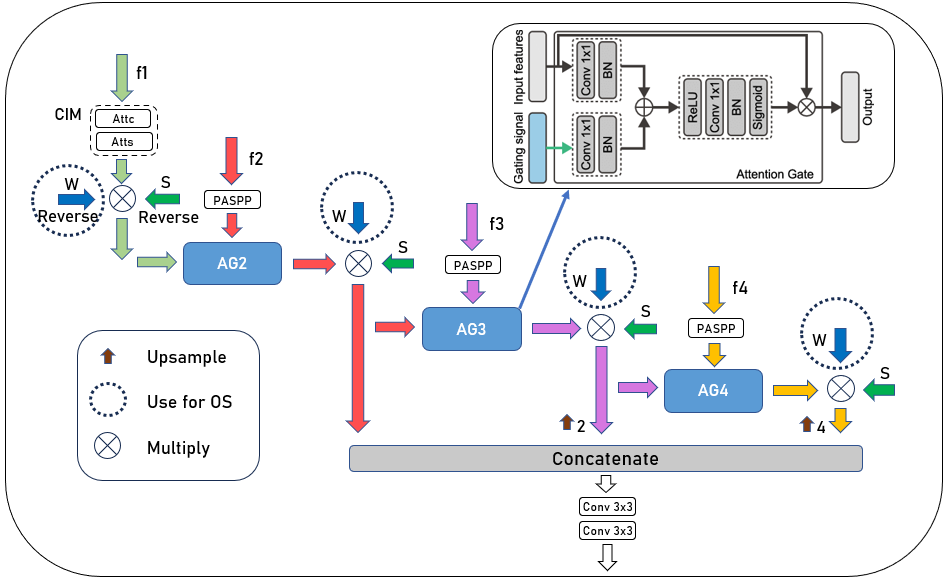}
\end{center}
\caption{Continuous Attention.}
\label{fig:refhinh3}
\end{figure}

\paragraph{Background Semantic.} At the early stage, the beginning mask rough prediction objects without clear structural details. To boost the segmentation performance, the details around the objects need to be detected and classified more precisely. Several previous works used reverse attention\cite{chen2018reverse} to highlight the outside area of objects to process the boundary constraint without correcting inaccurate details through texture information at low-level features. While low-level features often hold rich detail information, such as texture, color, edges, and polyps frequently resemble the background in appearance. Inspired by that motivation, we utilize $f_1$ to guide high-level features, modify inaccurate details around polyp objects and highlight strong components (S) to give a background semantic mask (BS).
\begin{align}
BS = Conv3x3(Conv3x3(Cat(S \otimes AG_i),i=2,3,4))
\end{align}
\paragraph{Object Semantic.} Recent architectures can not segment sufficiently recognized objects by creating a final mask only with an encoder and a decoder. Following our findings, several areas the model can detect, but they are so weak and easy to vanish through decoder progress because of the highlight of strong details, or detected areas but very noisy with the surrounding space. In addition, polyp colonoscopy images are one of the most difficult objects to segment because of the similarity between objects and backgrounds. Discovering weak features or challenging objects will have the potential to give outstanding performance. Therefore, instead of only using the strong (S) component as in the background semantic, we combine weak (W) and strong (S) components to explore and recover uncertain areas.
\begin{align}
OS = Conv3x3(Conv3x3(Cat((S,W) \otimes AG_i),i=2,3,4))
\end{align}
To implement the two ideas proposed above, we introduce Continuous Attention as depicted in~\autoref{fig:refhinh3} including gate attention mechanisms that are executed consecutively to modify the semantics of input high-level features $\{f_2,f_3,f_4\}$ before concatenating them to provide OS and BS as described in~\autoref{fig:refhinh2}. 

\subsection{Loss Function}
When analyzing polyp images, the consideration of boundary and area contributes greatly to segmentation performance, so we propose using a loss function described below. Whereas ACE loss~\cite{chen2020learning} offers advantages in terms of geometrical limitations, compact curvature, length, and area of active contours with region similarity, whereas BCE loss~\cite{xie2015holistically} represents a pixel restriction. We define the loss function as follows: 
\begin{equation}
Loss(y,\overline{y}) =  ACE(y,\overline{y})  + BCE(y,\overline{y}) 
\end{equation}

Where $\overline{y}$, $y$ denote the predicted mask and ground truth, respectively. 

\textbf{ACE Loss}
\begin{align}
ACE(y,\overline{y})= &( \alpha + \beta\overline{K}^2)|\nabla \overline{y}| 
+ \lambda y(c_1 - \overline{y})^2 \nonumber \\
&+ \lambda (1-y)(c_2 - \overline{y})^2 
\end{align}

Where $\alpha$ and $\beta$ execute the trade-off of length and curvature. $\lambda$ is used to balance two clauses. $\overline{K}$ is the curvature of $y$, $c_1$ and $c_2$ are mean intensities of the interior and outer regions.

\textbf{BCE Loss}
\begin{equation}
BCE(y,\overline{y})= \displaystyle \sum_{i=1}^{n} y_ilog(\overline{y_i}) + (1-y_i)log(1-\overline{y_i})  
\end{equation}

\section{Experiments}
This section describes the dataset, evaluation metrics, experimental results, and further insights into the output. From experimental results, we evaluate and compare ADSNet with current cutting-edge techniques for the polyp segmentation challenge. 

\paragraph{Dataset}
\begin{itemize}
 \item Kvasir-Seg~\cite{jha2020kvasir} There are 1000 polyp images and related annotations. The sizes range from 332x487 to 1920x1072, and the polyps that appear in the images likewise have a range in size and shape. 
    \item CVC-ClinicDB~\cite{bernal2015wm} consists 612 images with the size of 384 × 288 from 25 colonoscopy videos.
    \item  ETIS~\cite{bernal2012towards} contains 196 images with the size of 1225 × 966 collected from 34 colonoscopy videos. Because polyps are frequently small and challenging to locate, this dataset is more challenging.
    \item CVC-ColonDB~\cite{silva2014toward} chooses 380 images from 15 different colonoscopy sequences.
\end{itemize}

\paragraph{Evaluation Metrics}
We use some standard metrics for the purpose of segmenting medical images. The Dice Similarity Coefficient (DSC)~\cite{milletari2016v} statistic and Intersection over Union (IoU)~\cite{margolin2014evaluate} term are used to quantify the degree of similarity between two samples. MAE~\cite{margolin2014evaluate} shows the average absolute error between the true mask and the anticipated mask.  Weighted F-measure ($F^w_\beta$)~\cite{margolin2014evaluate} measures the effect of false negative, false positive, and true positive.

\subsection{Results}
We set up the same training and testing with other methods which are compared in the next section: 1450 (900 images from Kvasir-Seg~\cite{jha2020kvasir} and 550 images from CVC-ClinicDB~\cite{bernal2015wm}) for training. ETIS~\cite{bernal2012towards}, CVC-ColonDB~\cite{silva2014toward} for generalization ability testing, and remaining images in Kvasir-Seg~\cite{jha2020kvasir} and CVC-ClinicDB~\cite{bernal2015wm} for learning ability evaluation.

\begin{table}[htbp]
\begin{center}
\begin{tabular}{c|c|c}
\hline
\textbf{Methods}&\multicolumn{1}{|c|}{\textbf{Kvasir-Seg}}&\multicolumn{1}{|c}{\textbf{CVC-ClinicDB}} \\
\cline{2-3} 
\textbf{} & \textbf{\textit{Dice}}  \hspace{0.5cm} \textbf{\textit{IoU}} \hspace{0.5cm} \textbf{\textit{$F^w_\beta$}} \hspace{0.5cm} \textbf{\textit{MAE}} & \textbf{\textit{Dice}} \hspace{0.5cm} \textbf{\textit{IoU}} \hspace{0.5cm} \textbf{\textit{$F^w_\beta$}} \hspace{0.5cm} \textbf{\textit{MAE}} \\
\hline
DCRNet & \hspace{0.1cm} 0.886   \hspace{0.3cm}  0.825 \hspace{0.3cm}  0.868 \hspace{0.3cm}  0.035 & \hspace{0.1cm} 0.896  \hspace{0.3cm}  0.844  \hspace{0.3cm}  0.890 \hspace{0.3cm}  0.010\\
PraNet & \hspace{0.1cm} 0.898 \hspace{0.3cm}  0.840 \hspace{0.3cm}  0.885 \hspace{0.3cm}  0.030 & \hspace{0.1cm} 0.899 \hspace{0.3cm}  0.849 \hspace{0.3cm}   0.896 \hspace{0.3cm}  0.009\\
SANet & \hspace{0.1cm} 0.904  \hspace{0.3cm} 0.847 \hspace{0.3cm}   0.892 \hspace{0.3cm}  0.028 & \hspace{0.1cm} 0.916   \hspace{0.3cm}  0.859 \hspace{0.3cm}  0.909 \hspace{0.3cm}  0.012\\
Polyp-PVT & \hspace{0.1cm} 0.917   \hspace{0.3cm}  0.864 \hspace{0.3cm}  0.911 \hspace{0.3cm} 0.023 & \hspace{0.1cm} 0.937   \hspace{0.3cm}  0.889 \hspace{0.3cm}  0.936 \hspace{0.3cm}  \textbf{0.006}\\
Ours &  \hspace{0.1cm} \textbf{0.920}  \hspace{0.3cm} \textbf{0.871} \hspace{0.3cm}  \textbf{0.916} \hspace{0.3cm}  \textbf{0.020} & \hspace{0.1cm} \textbf{0.938} \hspace{0.3cm} \textbf{0.890} \hspace{0.3cm}  \textbf{0.940} \hspace{0.3cm}  \textbf{0.006}\\
\hline
\end{tabular}
\end{center}
\caption{Quantitative evaluation of diverse models on Kvasir-Seg \& CVC-ClinicDB Datasets.
The best results are bolded.}
\label{tab:tab1}
\end{table}

\begin{figure}[htp]
    \centering
     \includegraphics[width=.71\linewidth]{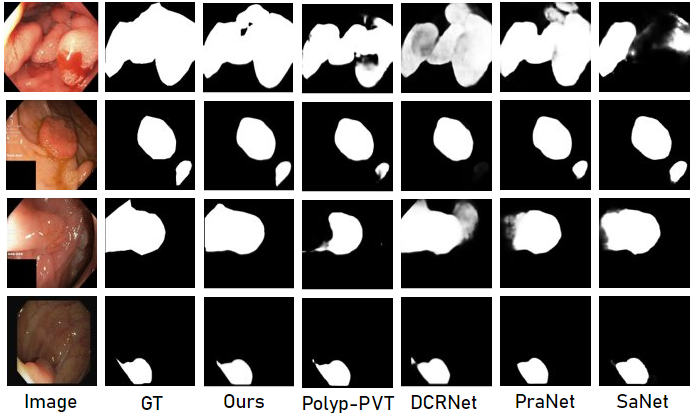}
    \caption{Qualitative analysis on Kvasir-Seg and CVC-ClinicDB dataset of different models in several noteworthy cases.}
    \label{fig:refhinh4}
\end{figure}

\paragraph{Learning ability.} In this experiment, the domain of the test and train set is similar. We compare ADSNet with recent SOTA methods including: PraNet~\cite{fan2020pranet}, DCRNet~\cite{yin2022duplex}, SANet~\cite{wei2021shallow}, and Polyp-PVT~\cite{dong2021polyp}. For a fair comparison, all results of the models mentioned above are referenced from the original papers. The detailed results shown in~\autoref{tab:tab1} indicate that ADSNet improves recent approaches. 0.920 Dice score and 0.871 IoU score on Kvasir-Seg~\cite{jha2020kvasir} and 0.938, 0.890 on CVC-ClinicDB~\cite{bernal2015wm}, respectively. Besides, our model also is better on $F^w_\beta$ and MAE indexes. For qualitative results,~\autoref{fig:refhinh4} shows obtained segmentation performance by all models on Kvasir-Seg~\cite{jha2020kvasir} and CVC-ClinicDB~\cite{bernal2015wm}. The segmentation performance of competitors is referenced completely from public sources in Polyp-PVT~\cite{dong2021polyp}. The great agreement between predicted samples by ADSNet and ground truths is illustrated. This proves the efficient and accurate segmentation ability of the proposed architecture in difficult and challenging situations with uncertain areas that previous approaches have not dealt with yet.

\begin{figure}[htp]
\begin{center}
 \includegraphics[width=.71\linewidth ]{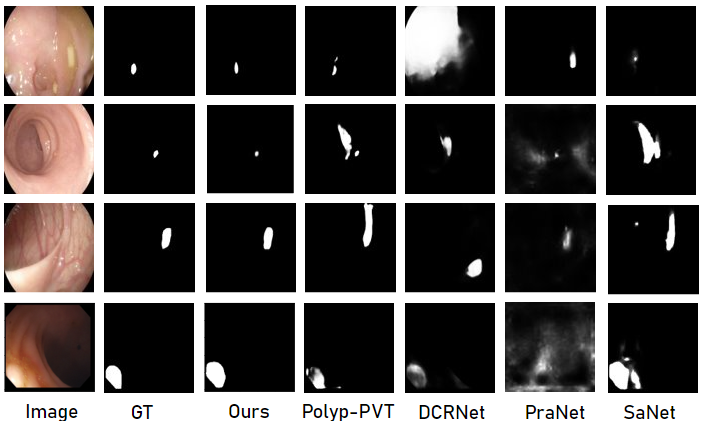}
\end{center}
\caption{Qualitative analysis on ETIS and CVC-ColonDB dataset of different models in several noteworthy cases.}
\label{fig:refhinh5}
\end{figure}

\noindent \textbf{Generalization ability.}
The compared results are shown in~\autoref{tab:tab2}. On ETIS~\cite{bernal2012towards} dataset, the performance obtain 0.798 Dice score and 0.715 IoU score improved by 1.39\% and 1.27\% over Polyp-PVT~\cite{dong2021polyp} respectively. ADSNet also shows great generalizability in CVC-ColonDB~\cite{silva2014toward} when surpassing all other SOTA methods with the highest scores, 0.815 Dice and 0.730 IoU score. Although on CVC-ColonDB~\cite{silva2014toward} the $F^w_\beta$ reduces when compared with SANet~\cite{wei2021shallow}, our proposal still improves on the remaining. Several noteworthy and difficult samples are shown in~\autoref{fig:refhinh5}. It is clear that ADSNet is capable of partitioning complex and minute details in obscure situations in addition to segmenting well in challenging cases with uncertain areas.   

\begin{table}[htbp]
\begin{center}
\begin{tabular}{c|c|c}
\hline
\textbf{Train Set}&\multicolumn{2}{|c}{\textbf{Kvasir-Seg \& CVC-ClinicDB}} \\
\cline{1-3} 
\textbf{Test Set} & \multicolumn{1}{c|}{\textbf{CVC-ColonDB}} & \multicolumn{1}{|c}{\textbf{ETIS}} \\

\cline{1-3} 
\textbf{Methods} & \textbf{\textit{Dice}}  \hspace{0.5cm} \textbf{\textit{IoU}} \hspace{0.5cm} \textbf{\textit{$F^w_\beta$}} \hspace{0.5cm} \textbf{\textit{MAE}} & \textbf{\textit{Dice}} \hspace{0.5cm} \textbf{\textit{IoU}} \hspace{0.5cm} \textbf{\textit{$F^w_\beta$}} \hspace{0.5cm} \textbf{\textit{MAE}} \\
\hline
DCRNet & \hspace{0.1cm} 0.704     \hspace{0.3cm}  0.631 \hspace{0.3cm}  0.684 \hspace{0.3cm}   0.052 & \hspace{0.1cm} 0.556    \hspace{0.3cm}  0.496  \hspace{0.3cm}  0.506 \hspace{0.3cm}  0.096\\

PraNet & \hspace{0.1cm} 0.712   \hspace{0.3cm}  0.640 \hspace{0.3cm}  0.699 \hspace{0.3cm}   0.043 & \hspace{0.1cm} 0.628   \hspace{0.3cm}  0.567 \hspace{0.3cm}  0.600 \hspace{0.3cm}  0.031\\

SANet & \hspace{0.1cm} 0.753    \hspace{0.3cm} 0.670 \hspace{0.3cm}   \textbf{0.892} \hspace{0.3cm}  0.043 & \hspace{0.1cm} 0.750     \hspace{0.3cm}  0.654 \hspace{0.3cm}  0.685 \hspace{0.3cm}  0.015\\
Polyp-PVT & \hspace{0.1cm} 0.808    \hspace{0.3cm}  0.727 \hspace{0.3cm}  0.795 \hspace{0.3cm} 0.031 & \hspace{0.1cm} 0.787     \hspace{0.3cm}  0.706 \hspace{0.3cm}  0.750 \hspace{0.3cm}  0.013\\

Ours & \hspace{0.1cm} \textbf{0.815} \hspace{0.3cm} \textbf{0.730} \hspace{0.3cm}  0.860 \hspace{0.3cm} \textbf{0.029} & \hspace{0.1cm} \textbf{0.798} \hspace{0.3cm} \textbf{0.715} \hspace{0.3cm}  \textbf{0.792} \hspace{0.3cm}  \textbf{0.012}\\
\hline
\end{tabular}
\end{center}
\caption{Quantitative evaluation of diverse models on unseen Datasets ETIS \& CVC-ColonDB.
The best results are bolded.}
\label{tab:tab2}
\end{table}

\begin{figure}[htp]
\begin{center}
 \includegraphics[scale=0.63]{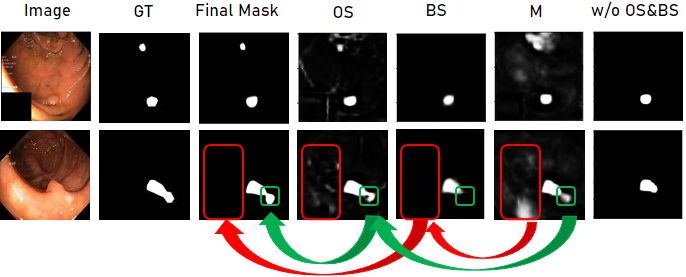}
\end{center}
\caption{The contribution of OS, BS, and M.}
\label{fig:refhinh6}
\end{figure}

\subsection{Further insights}
\begin{figure}[htp]
\begin{center}
 \includegraphics[scale=0.47]{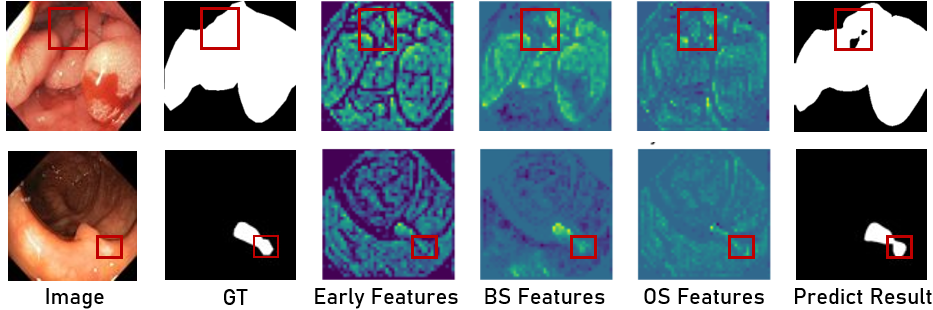}
\end{center}
\caption{Feature map of the early, BS and OS.}
\label{fig:refhinh7}
\end{figure}

This part proves the effectiveness of the combination of OS, BS, and M in the final mask. When the background semantic (BS) modifies wrong details around and highlights the primary structure of objects. The object semantic (OS) explores ambiguous and challenging areas. We display several segmentation performances in~\autoref{fig:refhinh6}. It can be seen that OS and BS assist the model capture context information in both object and background semantics. Meanwhile, without OS or BS, the architecture has difficulty in detecting and categorizing uncertain details, causing misclassification as in the last column. We also visualize the output feature map of the early, background semantic, and object semantic in~\autoref{fig:refhinh7} to verify the effectiveness of analyzing two distinct semantics, ADSNet offers significantly more sufficient features. 
\section{Conclusion}
In this paper, we have proposed a new novel architecture called ADSNet for exploring uncertain areas to improve polyp segmentation performance. We have introduced a new SOTA decoder that utilizes high-level features from the CNN-based encoder to produce an early global map. Finally, we have analyzed two separate semantics of the early global map: background semantic and object semantic using continuous attention to give an excellent map. The obtained results demonstrate the superiority of the proposed model in dealing with challenging cases with uncertain areas, from that, obtain better Dice, IoU, $f^w_\beta$, and MAE scores over recent state-of-the-art methods.

\paragraph{Acknowledgements} This work was supported by Institute of Information \& communications Technology Planning \& Evaluation (IITP) under the Artificial Intelligence Convergence Innovation Human Resources Development (IITP-2023-RS-2023-00256629) grant funded by the Korea government(MSIT). This work was also supported by the Basic Science Research Program through the National Research Foundation of Korea (NRF) funded by the Ministry of Education (NRF2021R1I1A3A04036408). This work was supported by Institute of Information \& communications Technology Planning \& Evaluation (IITP) grant funded by the Korea government(MSIT) (No.2021-0-02068, Artificial Intelligence Innovation Hub). The corresponding author is Soo-Hyung Kim.

\bibliography{main}

\end{document}